\documentclass{article}

\usepackage[preprint]{neurips_2023}

\usepackage[utf8]{inputenc} 
\usepackage[T1]{fontenc}    
\usepackage{microtype}
\usepackage{graphicx}
\usepackage{booktabs} 
\usepackage{algorithm}
\usepackage{algorithmic}
\usepackage{nicefrac}       
\usepackage{hyperref}
\usepackage{comment}
\usepackage{subcaption}
\usepackage{tabularx}
\usepackage{enumitem}
\newcounter{inlineenum}
\renewcommand{\theinlineenum}{\alph{inlineenum}}
\newenvironment{inlineenum}
  {\unskip\ignorespaces\setcounter{inlineenum}{0}%
   \renewcommand{\item}{\refstepcounter{inlineenum}{\textit{\theinlineenum})~}}}
  {\ignorespacesafterend}

\usepackage{amsmath}
\usepackage{amssymb}
\usepackage{mathtools}
\usepackage{amsthm}
\usepackage{enumitem}

\usepackage[capitalize,noabbrev]{cleveref}

\theoremstyle{plain}

\theoremstyle{definition}

\theoremstyle{remark}

\usepackage[textsize=tiny]{todonotes}
\usepackage{graphicx}

\title{Dynamic Interpretable Change Point Detection}

\author{%
  Kopal Garg\\
  \And
  Jennifer Yu\\
  \AND
  Tina Behrouzi\\
  \And
  Sana Tonekaboni\\
  \And
  Anna Goldenberg\\
}

\begin{document}

\maketitle

\newcommand{\ourCPD}{TiVaCPD}

\begin{abstract}
Identifying change points (CPs) in a time series is crucial to guide better decision making across various fields like finance and healthcare and facilitating timely responses to potential risks or opportunities. Existing Change Point Detection (CPD) methods have a limitation in tracking changes in the joint distribution of multidimensional features. In addition, they fail to generalize effectively within the same time series as different types of CPs may require different detection methods. As the volume of multidimensional time series continues to grow, capturing various types of complex CPs such as changes in the correlation structure of the time-series features has become essential. To overcome the limitations of existing methods, we propose TiVaCPD, an approach that uses a Time-Varying Graphical Lasso (TVGL) to identify changes in correlation patterns between multidimensional features over time, and combines that with an aggregate Kernel Maximum Mean Discrepancy (MMD) test to identify changes in the underlying statistical distributions of dynamic time windows with varying length. The MMD and TVGL scores are combined using a novel ensemble method based on similarity measures leveraging the power of both statistical tests. We evaluate the performance of TiVaCPD in identifying and characterizing various types of CPs and show that our method outperforms current state-of-the-art methods in real-world CPD datasets. We further demonstrate that TiVaCPD scores characterize the type of CPs and facilitate interpretation of change dynamics, offering  insights into real-life applications.
\end{abstract}

\section{Introduction}

In domains such as healthcare and finance, real-world time-series data is highly influenced by change points \citep{truong2018litreview}. Identifying and analyzing these changes in data distribution not only enhances our comprehension of the underlying patterns but also helps mitigate risks and improve decision-making.
 Change point detection (CPD) methods segment a time-series into distinct intervals with varying underlying properties. 
Precisely inferring time points associated with such transitions is essential to decipher the behaviors of the processes being modeled \citep{aminikhanghahi2017litreview}. With a substantial increase in the volume of time-series data collected in a variety of domains such as finance \citep{financecpd} and healthcare \citep{yang2006}, the importance of CPD methods that automatically capture changes in the signal has grown. CPD provides a way for the identification and localization of sudden changes in signals, such as patient vital signs, without the need for labeled information. For instance, in a medical setting, CPD enables the timely identification of significant variations such as changes in heart rate or declines in oxygen saturation levels, alerting doctors to potential health issues that demand immediate attention.
Most existing CPD methods fail to consider the underlying variability in the properties and root causes of change points (CPs) and therefore cannot generalize effectively to time-series with complex change dynamics. CPs can be characterized as changes in the distribution of the measurements over time. Alternatively, they could also be the result of changes in the correlation structure between features. The former is more studied in the literature, but the latter is also of significance in many applications. 
For instance, among physiological signals, the Heart Rate Variability (HRV) measure always shows a negative correlation with the Heart Rate (HR) measurement, but in some rare situations, they might exhibit a positive correlation, indicating a concerning change in the underlying health state of an individual. At the same time, changes in the average HR can be indicative of other conditions. Methods that rely solely on detecting changes in marginal distributions will fail to identify such scenarios.

\begin{figure*}[ht!]
    \centering
    \includegraphics[width=.7\textwidth]{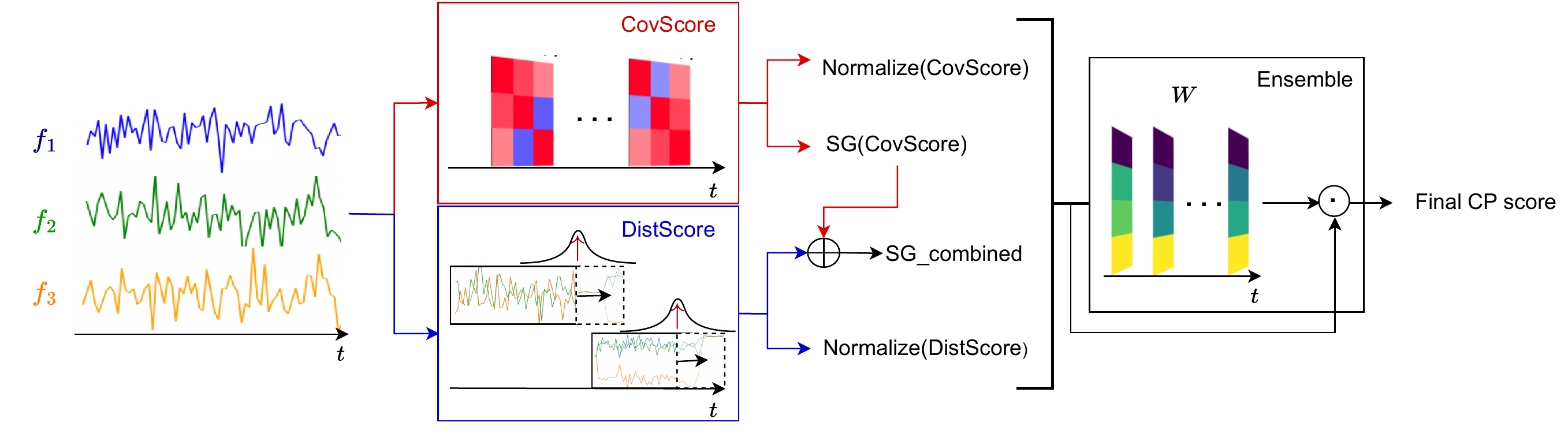}
    \caption{\ourCPD\ overview that shows an abstract illustration of the generation process for DistScore, CovScore, and the ensemble weights $W$ used to aggregate the components of the model. SG stands for Savitzky-Golay filter.} 
    \label{fig:overview}
\end{figure*}

In this paper, we propose a statistical CPD scoring method called \ourCPD\ that captures different types of CPs in time-series without the need for labeled instances of change. \ourCPD\ offers a non-parametric solution that does not require any distributional assumption of the generative process, and can therefore generalize to various scenarios. 
TiVaCPD, as shown on Figure \ref{fig:overview}, assigns a CP score to each time-point that consists of two parts: 1) change in the correlation of features and 2) change in the underlying distribution of the time-series features. Each part of the score is interpretable, allowing us to characterize and classify CPs with similar underlying properties and understand the cause of the change. 
To identify changes in the joint distribution of features over time, we rely on the hypothesis that changes in feature interactions can be effectively captured via correlation networks constructed from adjacent windows of time. To this end, we employ a dynamic network inference method, time-varying graphical lasso (TVGL) \citep{hallac2017tvgl}, to acquire sparse time-varying precision matrices to detect changes in feature correlation patterns. 
To identify changes in the probability distributions of adjacent windows, we build on recent statistical results by \citet{schrabb2021mmdagg} in the theory of non-parametric two-sample MMD tests. 
Unlike existing CPD methods that employ a fixed-size sliding-window approach, we dynamically establish the window size, accounting for the variable length of states between CPs. Fixing window sizes introduces issues: small windows are likely to compromise the power of the statistical test, and larger windows are at risk of aggregating different distributions. By dynamically adjusting the sliding window size, we address these concerns. 
We propose to ensemble different components of the TiVaCPD score. Briefly, the ensemble method adaptively assigns weights to scores based on their dissimilarity, placing greater emphasis on scores that capture changes not detected by other components.
We evaluate our method’s ability in identifying various categories of CPs on 4 simulated and 2 real-life time-series datasets and compare its performance against 3 state-of-the-art CPD methods; showing that our method outperforms all competitors across real-word and one complex simulated datasets. There are four main contributions of this work:
\vspace{-2mm}
\begin{itemize}[noitemsep,topsep=6pt, leftmargin=*]
    \item We detect and characterize different types of changes in feature dynamics and/or distribution. The categorization and visualization of the underlying CP cause enhances interpretability providing valuable insights of the observed patterns. 
    \item We present a novel use of Time-Varying Graphical Lasso for quantifying changes in feature interactions. We demonstrate its ability to detect CPs that occur due to changes in the covariance of the joint distribution of features over time. 
    \item We introduced a dynamic window selection method that effectively addresses the limitations of static windows when detecting changes in data distribution.  
    \item We propose a novel ensemble technique to best aggregate unsupervised CP scores from different statistical tests. We also introduce a post-processing procedure to smooth the score estimate while preserving local minima and maxima, using the Savitsky-Golay filter \citep{sgfilter}. 
\end{itemize}

\vspace{-4mm}
\section{Related Work}
There is abundant literature on CPD methods  \citep{truong2018litreview, aminikhanghahi2017litreview, reeves2007}. CPD methods consider a time-series to be a collection of random variables with abrupt changes in distributional properties over time. Most of these methods are parametric \citep{yamanishi2002, kawahara2007} and involve estimating the underlying probability density function of the signal, which limits detection to certain type of distributions and is usually computationally expensive. 
Non-parametric methods \citep{chang2018kernel,cheng2020optimal,matteson2013} are used where the time-series dynamics cannot be easily modeled and prior assumptions about the data distribution cannot be made. An optimal transport-based method proposed by \cite{cheng2020optimal}, conducts two-sample Wasserstein tests between the cumulative distribution of contiguous subsequences. It uses fixed-size sliding windows to compute the test statistic. However, basing CP decisions on the local maxima of this statistic can result in a higher false positive rate. Moreover, this method projects the data onto one-dimension and uses the mean statistic, potentially leading to the loss of detection power. 

Deep learning-based methods are another type of non-parametric approach that recently gained popularity due to the increasing amount of available data. For example, Time-Invariant Representation (TIRE) \citep{tire}, an autoencoder-based CPD approach, learns a partially time-invariant representation of time series and computes CPs using a dissimilarity measure. Another deep learning technique, referred to as $TS-CP^{2}$ \citep{tscp2}, utilizes contrastive learning to detect CPs, leveraging the representation of time series acquired from temporal convolutional networks. Other deep learning-based CPD methods use kernel functions \citep{li2015} for greater flexibility in representing the density functions of intervals of time. One such method, KLCPD \citep{chang2018kernel} uses deep generative models to increase the test power of the kernel two-sample MMD test statistic \citep{gretton2007}. It overcomes limitations of prior kernel-based CPD methods by removing the need of a fixed number of CPs or relying on prior knowledge of a reference or training set for kernel calibration. However, its performance depends on the choice of kernel and kernel bandwidths. 

The lack of interpretability in deep learning-based CPD methods hinders our understanding of why and how these methods make predictions. Moreover, current methods often fail to capture changes in correlation patterns that occur due to evolving dynamics of multivariate time series. To address such CPs, \citet{gibberd2017graphtime} introduced GraphTime; A Group-Fused Graphical Lasso estimator for grouped estimation of CPs in dependency structures of a time-series captured by a dynamic Gaussian Graphical Model. As the estimated graph topology is piece-wise constant, this is useful only when we are interested in detecting jump points and abrupt changes, and leads to excessive number of false positives for other gradual CPs. Another CPD method \citep{RoyAtchade} looks for dependencies between spatial or temporal variables using high-dimensional Markov random-field model. This method relies on two assumptions of  known covariance structure and stationary for the data. These are a strong assuption that might not hold in many real-world applications.



\vspace{-2mm}
\section{Method}

\paragraph{Problem Formulation}
Consider a multivariate time-series sample $X \in \mathbb{R}^{d\times T}$ to be a sequence of random variables $[X_1, X_2, ..., X_T]$ with $d$ indicating the number of features in $X$. $T$ represents the total number of measurements over time. 
To identify change points in time steps of a data sample $X$, a score $S[t]$, $\forall t \in [T]$ is estimated for each time step that measures the amount of change in the underlying generative distribution of the data. 
\paragraph{Our CPD Algorithm - \ourCPD}
In this section, we introduce our CPD algorithm called Time Variable Change Point Detection (\ourCPD).
The score $S$ generated by \ourCPD\ is composed of two components: 1) a score that measures change in correlation-CovScore (Algorithm \ref{alg:corr_score}), 2) a score that measures change in the distribution-DistScore (Algorithm \ref{alg:dynamic_windowing}). In the rest of the section we introduce each component separately, explain how each results in a unifying score that captures a variety of CP types, and demonstrate how to interpret the score to better understand the CPs.

\subsection{ \bf Detecting changes in the correlation structure (CovScore)}

\begin{figure*}[ht!]
    \centering
    \begin{subfigure}[b]{0.55\textwidth}
    \centering
        \includegraphics[scale=0.65]{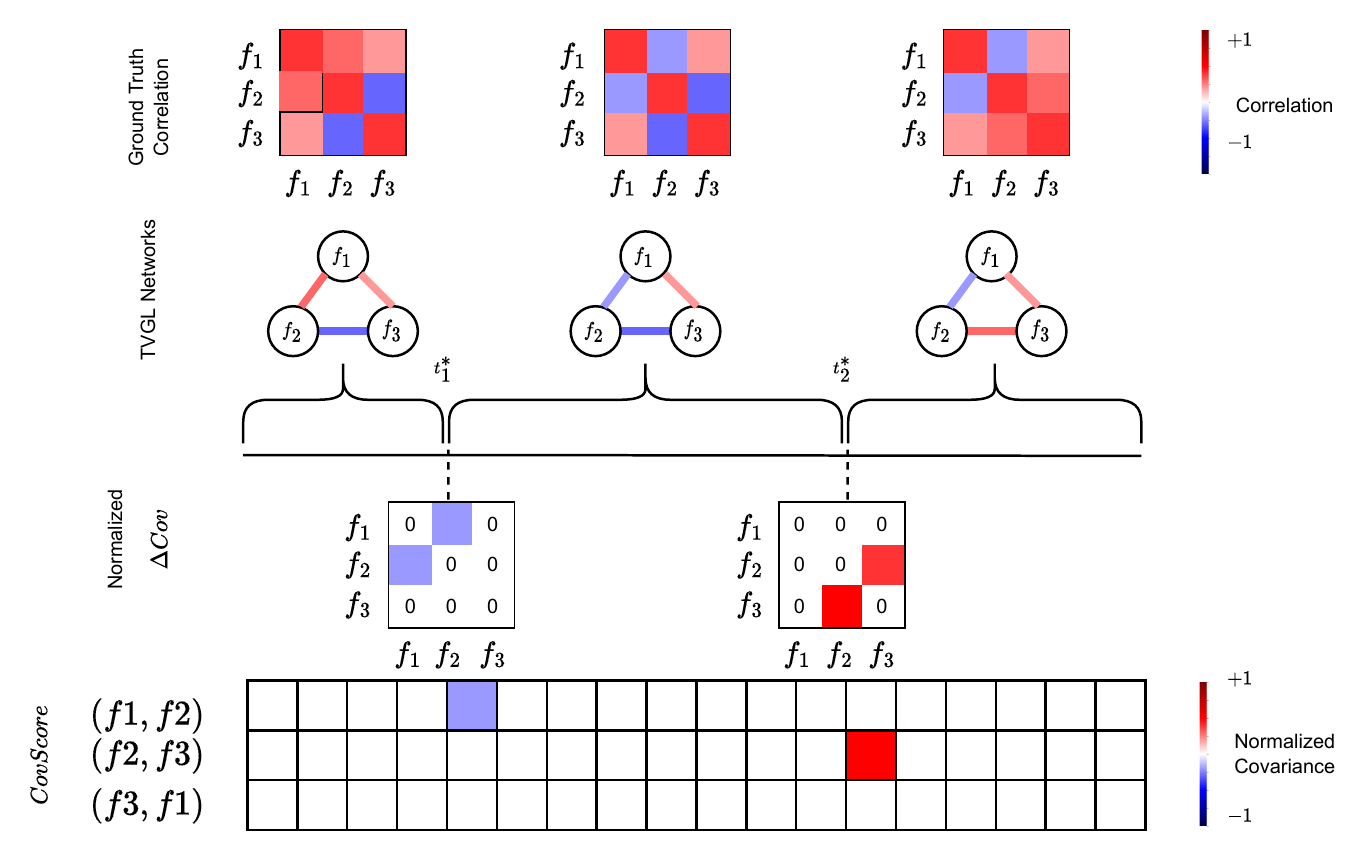}
        \caption{TVGL: This example shows changes in three feature interactions over time via a correlation matrix heatmap. Each row indicates the correlation shifts between feature pairs, with colours (blue or red) showing changing direction and colour intensity indicating magnitude.}
        \label{fig:tvgl}
    \end{subfigure}
    \hfill
     \begin{subfigure}[b]{0.35\textwidth}
     \centering
         {\includegraphics[scale=0.45, trim={0 0 0 0.25cm},clip]{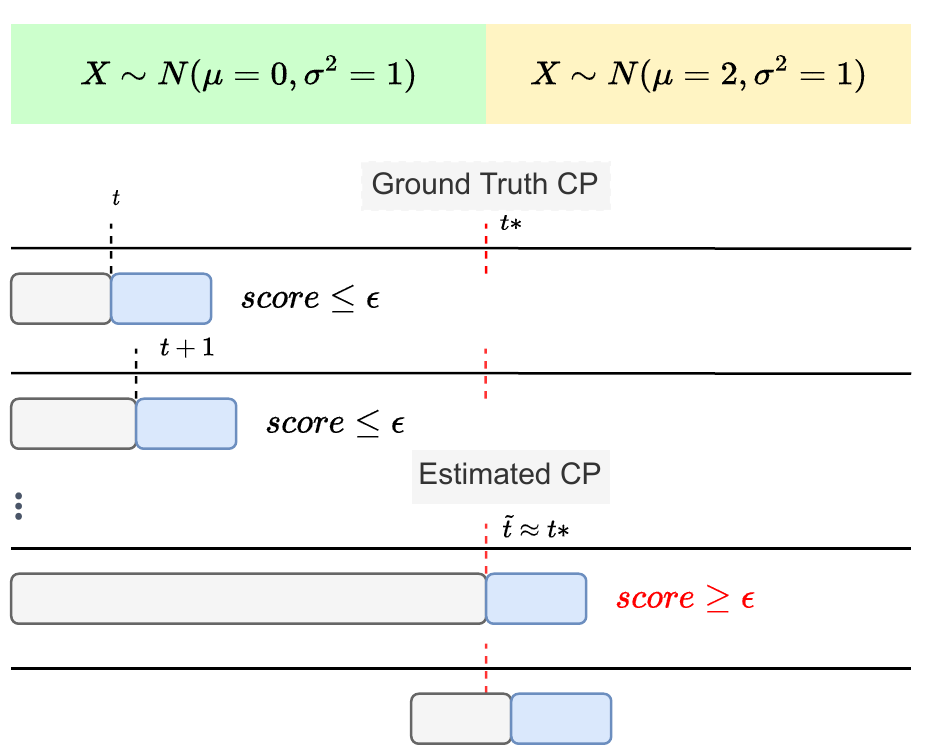}}
         \caption{Dynamic window: The expanding window (gray) and fixed-size future observation window (blue) enlarge to include more samples from the generative distribution as the algorithm proceeds. Once a CP is detected at $\widetilde{t}$, the window size reverts to its initial size.}
        \label{fig:windowing}
     \end{subfigure}
     \caption{TiVaCPD method breakdown }
\end{figure*}

A CP can be caused by a change in the correlation between features. This results in a change in the covariance of the joint distribution that can be identified as a state change in the feature network. The evolving dynamics of features can be modeled using graphical models, i.e. at every time point, the interactions of features can be modeled as a graph network, with nodes and links corresponding to each feature and correlation between sets of variables, respectively (Figure \ref{fig:tvgl}).
 However, in a time-series setting, estimating the graphical network at every time step is computationally challenging. 
To estimate these networks to detect CPs, we use Time-Varying Graphical Lasso (TVGL) \citep{hallac2017tvgl}, an efficient algorithm for estimating the inverse covariance matrix (precision matrix) of multivariate time-series with time-varying structures. TVGL infers the structure of graph networks by estimating a sparse time-varying inverse covariance matrix, $P_t = \Sigma^{-1}_t$ ($\Sigma_t$ indicating the covariance matrix at time $t$), of the variables for all $t\in[T]$. 
TVGL extends the graphical lasso problem to dynamic networks by allowing covariance $\Sigma$ to vary over time, taking into account how the relationships between signals evolve. 
TVGL method enforces sparsity in the covariance matrix for less computational cost and easier interpretability of the graphical network structure. 
A scalable message-passing algorithm called Alternating Direction Method of Multipliers \citep{admm} is employed to estimate the sparse inverse covariance. Additional details on the TVGL method can be found in the Supplementary Material.

Learning the inverse covariance estimates from windows of data reveals the underlying evolutionary patterns present in the time-series. We employ TVGL technique to identify points of change in feature interactions between adjacent local windows of time. 
To promote the identification of shifts in the covariance pattern of features, we integrated an L2-norm penalty function into the estimation of the matrix inverse. Moreover, to ensure the invertibility of $\Sigma_t$, we applied feature standardization and removed highly correlated features before feeding them to TVGL.
The partial correlation of two features can be estimated from the joint precision matrix entries as $-\frac{P(i,j)}{\sqrt{P(i,i)P(j,j)}}$, which means contrasting consecutive precision matrix entries over time quantifies the change in the features' correlation. 
By taking the difference between the absolute values of adjacent precision matrices, we quantify these changes in the distribution. A value close to 0 in this matrix indicates nearly identical estimations of the network and therefore no CP in that feature interaction (see Algorithm \ref{alg:corr_score}). A negative value indicates an increase in absolute correlation and a positive value indicates a decrease.

\begin{algorithm}[t]
\caption{Estimating CovScore}
\label{alg:corr_score}
\begin{algorithmic}[1]
 \STATE {\textbf{Input: $X$} (multivariate time-series)} 
 \STATE {\textbf{Output: $CovScore$} (score representing change in adjacent precision matrices)}
 \STATE {{Pre-processing:} Remove features with correlation higher that $0.95$}
    \STATE {$P = $TVGL$(X)  \quad $// Estimate the sparse inverse covariance $\forall t \in [1,\ldots,T]$ }
    \FORALL{$t \in [1,...,T]$}  
        \STATE CovScore[t] $= \left\{\begin{array}{cl}
           \sum (\lvert P[t]\rvert-\lvert P[t-1] \rvert) & \text{if $t>0$} \\
           0               & \text{if $t=0$}
         \end{array}\right.$ 
    \ENDFOR
    \STATE{{return} CovScore}
\end{algorithmic}
\end{algorithm}{}



\subsection{\bf Detecting shift in distribution (DistScore)}

Assuming in a time-series sample $X$, each $X_t$ is independently generated from a joint probability distribution $p_t(\cdot)$, a CP occurs at time $t^*$ if observations after $t^*$ are generated from a different distribution.
To compare the probability distributions of adjacent windows, we employ a non-parametric two-sample testing procedure called MMD Aggregate (MMDAgg), introduced in \citet{schrabb2021mmdagg}. 
Let $\Delta^-$ represent the initial size of the window of $X$ before a query point $t$, and let this prior window be denoted by $X_{\Delta^-}^t =\{X_{t-\Delta^-},...,X_{t-1},X_t\}$. Similarly, the window of future observations can be denoted by $X_{\Delta^+}^t =$ $\{X_{t+1},...,X_{t+\Delta^+-1},X_{t+\Delta^+}\}$, where ${\Delta^+}$ represents the length of a future window. Kernel-based MMD tests serve as a measure between two probability distributions. With the statistical test threshold $\alpha$, if the null hypothesis $H_0$, is rejected, the time-series may be partitioned by a CP at $t^*$, signifying that measurements in the $X_{{t^*}-\Delta^-:t^*}$ windows come from  a different distribution than measurements in $X_{t^*:t^*+\Delta^+}$.
The performance of a single kernel-based MMD test typically depends on the choice of kernel and kernel bandwidths. 
Since we compare adjacent windows with restricted number of samples, any loss of data to kernel bandwidth selection can be detrimental to our method's performance. To overcome this problem, MMDAgg aggregates multiple MMD tests using different kernel bandwidths, ensuring maximized test power over the collection of kernels used and eliminating the need for data splitting or arbitrary kernel selection. The method considers a finite collection of bandwidths where the aggregated test is defined as a test that rejects $H_0$ if one of the tests over a given bandwidth rejects $H_0$.

We propose to dynamically establish the window size based on the presence of CPs (Algorithm \ref{alg:dynamic_windowing}). Let $\Delta^-$ represent the size of the dynamic window of data points from the last estimated CP ($\widetilde{t}$), up until the current time point ($t$). Starting with a constant $\Delta^+$ and a small $\Delta^-$ window, the length of the running window $\Delta$ increases with each new observation until a new CP occurs, according to the MMD test. If a significant change in distribution isn't detected by the MMD test, i.e. the MMD score is smaller than a pre-defined threshold $\epsilon$, the two sub-sequences are combined and compared against the next sub-sequence in the series. 
This process is also explained in Figure \ref{fig:windowing}. Our dynamic windowing method eliminates the need for repetitive fixed-window comparisons and utilizes a growing sample set for the MMD test. 

For determining the final CP score, we need to meaningfully ensemble the  DistScore with the CovScore, which is challenging because the covariance score is bounded while MMD is a positive unbounded score. Hence, \ourCPD\ incorporates kernel normalization in the MMDAgg algorithm. 
we use a generalization of Cosine normalization \citep{ah2010normalized} to normalize our kernels so as to have a similarity index. For a given kernel function, $K^{z=1}(x, y)$ represents the normalized kernel of order $z$. We use the generalized mean with exponent $z=1$ (arithmetic mean),
which means $K^{z=1}(x,y) = \frac{K(x,y)}{M^{z=1}(K(x,x), K(y,y))}$, where $M^{z=1}(a_{i=1}^{p}) = \frac{1}{p}\sum^p_{i=1}(a^z_i)$.
This normalization technique projects the objects from the feature space to a unit hypersphere and guarantees $|K^{z=1}(x, y)| \leq 1$.

\begin{algorithm}[t]
\caption{Estimating DistScore using Dynamic MMD}
\label{alg:dynamic_windowing}
\begin{algorithmic}[1]
 \STATE {\textbf{Input: time-series $X$, $\alpha$ (Statistical test threshold), $\Delta^-$ (Initial previous window size), $\Delta^+$ (Future window size), $\epsilon$ (Threshold for significance)}}
 \STATE {\textbf{Output: DistScore}}
  \STATE{$\Delta = \Delta^- \quad $// Size of the running window}
  \FORALL{$t \in [1,...,T]$}
  \STATE{$S[t]$ = MMDAgg($X_{t-\Delta:t}$, $X_{t:t+\Delta^+}$, $\alpha$)}
   \IF{$S[t]$ $\geq \epsilon$ }
     \STATE {$DistScore[t]=S[t]$} \hspace{.5cm}\&\hspace{.5cm}  {$\Delta = \Delta^-$}
   \ELSE 
     \STATE{$DistScore[t]=0$} \hspace{.5cm}\&\hspace{.5cm} {$\Delta = \Delta + 1$}
   \ENDIF
  \ENDFOR
  \STATE{return DistScore}
\end{algorithmic}
\end{algorithm}{}


\vspace{-5pt}
\subsection{Ensemble Score and CP Categorization}

An effective method for combining unsupervised CPD scores is poorly studied. Here, we use an ensemble method that utilizes the score differences to highlight the scores that contribute the most to representing the CPs. Algorithm \ref{alg:overall} indicates our ensemble technique for generating a unified score $S$ by combing CovScore and DistScore. We use four score variants derived from CovScore and DistScore to generate dynamic dissimilarity weight $W$ to effectively aggregate the scores into a unified one (As shown in Figure \ref{fig:overview}). 
The four scores used are as follows: 
\begin{inlineenum}
\item \item Normalize(CovScore) and Normalize(DistScore): standardized CP scores using z-score normalization to bring them into the same scale,
\item SG(CovScore): Since CovScore is sensitive to small distributional changes that can lead to false change point detection, we mitigate the risk of detecting spurious CPs by applying Savitzky-Golay (SG) smoothing filter. SG filtering technique is a widely used method for smoothing patterns and reducing noise in time series data \citep{sgfilter}.
\item SG\_combined: We also apply the filter to the sum of filtered CovScore and DistScore, which effectively reduces noise and improves the performance of CPD. 
\end{inlineenum}
The ensemble approach utilizes a weighted average of aforementioned four scores.
The importance weight $W$ is determined by computing the mean absolute difference between scores. By assigning greater weight to scores that are more dissimilar, we can effectively identify CPs that may have been missed by other scoring methods. The weights are calculated for each 21 time-points window, as the distribution of scores' importance may vary over time.
To locate the exact time of the CPs, we look for peaks in the ensemble scores by searching for local maxima with a threshold that is used to remove number of false positive CPs created by noise. Using a threshold for peak detection is a common practice that requires careful tuning based on circumstances.

\subsection{Understanding the change points and interpreting \ourCPD\ score}
\begin{figure*}[ht!]
    \centering
    \includegraphics[width=.95\textwidth]{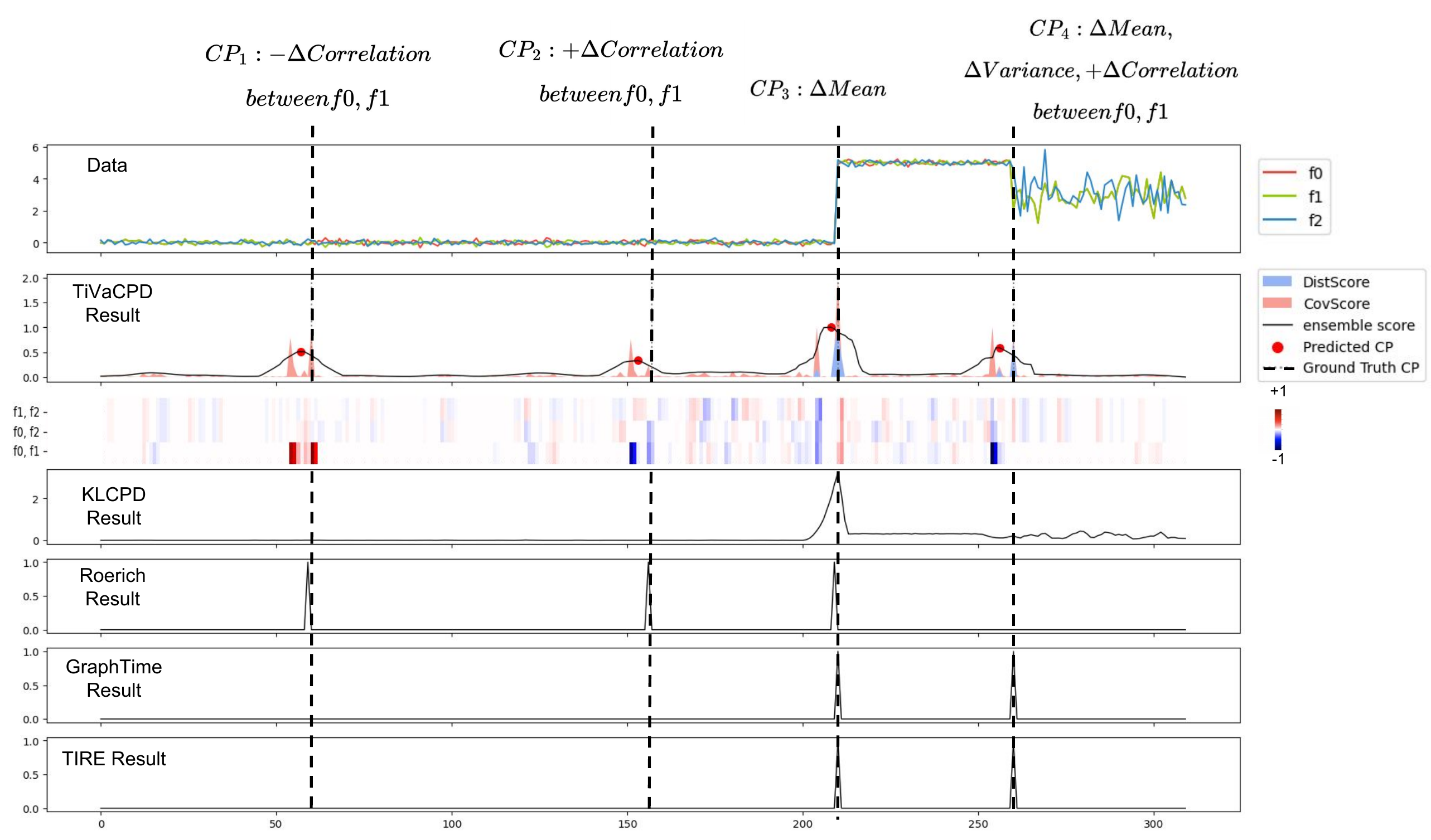}
    \caption{This example shows a simulated time-series with various CPs. Row 1: raw data. Rows 2-3: TiVaCPD score with heatmap interpretation. Rows 4-6: other CP baseline scores.}
    \label{fig:interactions_real_data}
\end{figure*}

\ourCPD \space offers valuable insights into the underlying nature of the observed change points. 
Detecting both changes in correlation and data distribution, such as an inverse correlation between heart rate variability and heart rate, is crucial in clinical practice, as it can indicate an adverse health event. The DistScore detects changes in the underlying distribution of the time-series, while the CovScore detects changes in the correlation structure of feature pairs, providing a detailed analysis of the feature dynamics at each time step. This information allows for the categorization of CPs, thereby enhancing interpretability, as shown in Figure \ref{fig:interactions_real_data}. This figure presents a multivariate time series sample showcasing TiVaCPD results, featuring ConvScore, DistScore, and a weighted ensemble score. In addition, CovScore's heatmap illustrates the feature pairs that caused a CP, and TiVaCPD identifies the direction of correlation change at each CP. It also includes a comparative analysis with other CPD methods, such as KLCPD, Roerich, GraphTime, and TIRE, providing a comprehensive evaluation of the TiVaCPD results against alternative CPD methods. The first two CPs ($CP_1$ and $CP_2$) are caused by changes in the correlation between features 0 and 1, where $CP_1$ corresponds to a negative change in correlation and $CP_2$ corresponds to a positive change in correlation. CP 3 is caused by changes in the mean between features 0 and 1 as the DistScore is high. CP 4 is caused by changes in a combination of variance, correlation, and mean, which are simulated to resemble real-world data.


\begin{algorithm}[t]
\caption{{\ourCPD}}
\label{alg:overall}
\begin{algorithmic}[1]
 \STATE {\textbf{Input: $X$}, $\alpha$, $\Delta^-$, $\Delta^+$, $\theta$ (Peak finding threshold)}
 \STATE{\textbf{Output: $CP$ (time points of CP), $S$ (TiVaCPD score) }}
\STATE {CovScore = TVCov$(X)$}
\STATE {DistScore = DynamicMMD$(X , \Delta^-, \Delta^+$, $\alpha$)} 
\STATE SG\_combined = SG($|$SG(CovScore)$|$+$|$DistScore$|$)
\STATE {All\_Score = [Normalize(DistScore), Normalize(CovScore),SG(CovScore), SG\_combined]}
 \FORALL{$t\_win \in [T]$}
    \STATE {$W = \sum_{j} mean(|\text{All\_Score} - \text{All
    \_Score}_j|)$} 
    \STATE $S = W  \boldsymbol{\cdot}  \text{ All\_Score}$
  \ENDFOR
 \STATE {$CP = $ PeakFinding$(S, \theta)$}
 \STATE{return $CP, S$}

\end{algorithmic}
\end{algorithm}{}
\section{Experiments}

\subsection{Datasets and Hyper-parameter Settings}

We demonstrate the performance of our method compared to multiple baselines on four simulated and two real-life multivariate datasets commonly used in the CPD literature. Different simulations test the functionality of CPD methods on a variety of potential CP scenarios. All hyper-parameters are determined based on random search over 10\% of the datasets (more details on best parameters and sensitivity to hyperparameter change are provided in the supplementary material). 

\paragraph{Simulated Data:} We created 4 different datasets to simulate different types of CPs. In all datasets, each time series sample $X \in {\mathbb R}^{d \times T}$ consists of $d=3$ features, and each $X_t$ is sampled independently from a Gaussian distribution $x_{i,t} \sim N(\mu_{i,t}, \sigma^2_{i,t})$.

\begin{itemize}[leftmargin=*,noitemsep,nolistsep]
\setlength\itemsep{-.75em}
\item \textbf{Jumping Mean}: For this dataset, the variance is assumed to be constant over time and across all features and is set to $\sigma^2=0.5$. The ground-truth CPs correspond to abrupt jumps in the mean that can happen independently in any of the features.\\
\item \textbf{Changing Variance}: In this dataset, all three features are generated with constant mean $\mu=1$, but their distribution variance changes over time. CPs are indicated as time points with changes in $\sigma^2$. \\
\item \textbf{Changing Correlation}: This data set consists of a multivariate time-series generated with constant $\sigma^2$ and $\mu$. To introduce correlation changes between variables, feature 2 is modified to be: $\widetilde{x_{2,t}}=\rho_t \times x_{1,t}+\sqrt{(1-\rho^2_t)} \times x_{2,t}$, where $\rho_t$ controls the correlation between the two features that can vary over time and are randomly sampled from $[-1,1]$. Here, the ground truth CPs correspond to points in time where the correlation $\rho_t$ changes.\\
\item \textbf{ Arbitrary CPs}:This data set consists of a multivariate time series with CPs due to varying $\mu$, or $\sigma^2$, or correlations between pairs of variables, resulting in a mixture of CPs scattered over time.
\end{itemize}
\paragraph{Real-world Data:}
\begin{itemize}[leftmargin=*,noitemsep,nolistsep]
\setlength\itemsep{-.75em}
\item \textbf{Bee dance} \citep{beedance2008}: This dataset consists of six three dimensional time-series of bees' positions while performing three-stage waggle dances. The bees communicate through actions like left/right turn, and waggle. The transition between the states represents ground truth CP.\\
\item \textbf{Human Activity Recognition (HAR)} \citep{har2013}: We use a subset of HAR which includes periods of 6 activities such as normal walking and standing. These activities are measured with 3-axial linear acceleration and angular velocity sensors, for a total of 6 features. The ground-truth CPs are labeled as the transitions between activities.
\end{itemize}

\subsection{Baseline Methods}
We compare the performance of \ourCPD\ with SOTA CPD methods on various types of CPs. \footnote{The implementation of the baselines was done using publicly released source code by the authors.} The selected SOTA approaches include those that measure a change in distribution and those that focus on the graphical structure of features over time. In addition, we conducted an ablation study on various parts of the \ourCPD\ score to analyse each component impact.

\textbf{Kernel Change Point Detection (KLCPD) \citep{chang2018kernel}} is a kernel learning framework for CPD that uses a two-sample test and optimizes a lower bound of test power via an auxiliary generative model. For this method, we used window sizes $w\in[10,25]$ for all experiments and trained the model for $25$ epochs, unless more training led to improved results. Consistent with our own post-processing steps, we performed peak detection to detect the exact time of change. 

\textbf{Roerich \citep{hushchyn2021generalization}} is a CPD method based on direct density ratio estimation to detect the change in distribution. We set all parameters to default and use window sizes $w\in[10,25]$ for all experiments.
      
\textbf{Group Fused Graph Lasso (GraphTime) \citep{gibberd2017graphtime}} is a time-varying graphical model based on the group fused-lasso. Similar to \ourCPD\, it uses the graphical model to model the dependencies of variables in time series. GraphTime models the temporal dependencies between variables while \ourCPD\ models the pairwise dependencies to identify CPs. 

\textbf{Time-Invariant Representation (TIRE) \cite{tire}} is an autoencoder-based CPD method, we used window sizes $w$ between ${10,25}$ for all experiments. For the $domain$ parameter, we chose $both$ to include both time, and frequency domains. We used $200$ epochs to train the model. 




\begin{figure} 
  \begin{minipage}[c]{0.5\textwidth}
    \includegraphics[scale=0.7]{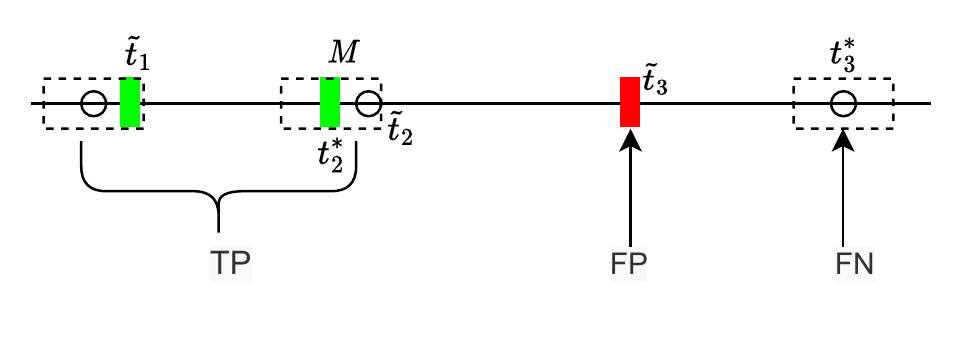}
  \end{minipage}\hfill
  \begin{minipage}[c]{0.5\textwidth}
    \caption{Definition of margin of error. The ground truth CPs ($t^*$) are shown as circles and $\widetilde{t}$ are the detected CPs. The prediction is a true positive if $\widetilde{t}$ falls within the margin around the ground truth.} \label{fig:03-03}
    \label{fig:evaluations}
  \end{minipage}
\end{figure}

\subsection{Evaluation} 
\begin{table*}[ht!]
\small
    \centering
    \begin{tabularx}{\textwidth}{lcccccc}
    & \multicolumn{3}{c}{Jumping Mean} &  \multicolumn{3}{c}{Changing Variance} \\
    \toprule
       Method  &  Precision & Recall & F1 (M=5) &  Precision & Recall & F1 (M=5) \\
       \toprule
       \ourCPD &1.00 (0.00)  &1.00 (0.00) & \textbf{1.00 (0.00)}  &0.86 (0.14)&0.73 (0.18) &{0.78 (0.15)} \\
       \ourCPD-CovScore &0.00 (0.00)  &0.00 (0.00) &0.00 (0.00)  &0.03 (0.07)&0.03 (0.06) &0.03 (0.06) \\
       \ourCPD-DistScore &1.00 (0.00)  &1.00 (0.00) & \textbf{1.00 (0.00)}  &0.79 (0.14)& 0.75 (0.17)&0.76 (0.15)\\
       \midrule
       KL-CPD &0.76 (0.11)  &0.83 (0.09) &0.79 (0.09)  &0.18 (0.10) &0.30 (0.20) &0.20 (0.15)\\
       Roerich & 1.00 (0.00) &1.00 (0.00) & \textbf{1.00 (0.00)}  & 1.00 (0.00) &1.00 (0.00) & \textbf{1.00 (0.00)}\\
      \midrule
    GraphTime  &1.00 (0.00)  &1.00 (0.00) & \textbf{1.00 (0.00)}  &0.05 (0.01) &1.00 (0.00) &0.09 (0.02)\\ 
    TIRE &0.90 (0.23)  &0.73 (0.29) &0.77 (0.26)  &0.78 (0.24) &0.60 (0.23) &0.66 (0.22)\\ 
    \end{tabularx}
    \caption{Performance of CPD methods on the Jumping Mean and Changing Variance with $M=5$.}
    \label{table:jmcv5}
\end{table*}


\begin{table*}[ht!]
\small
    \centering
    \begin{tabularx}{\textwidth}{lcccccc}
    & \multicolumn{3}{c}{Changing Correlations} &  \multicolumn{3}{c}{Arbitrary CPs} \\
    \toprule
       Method  &  Precision & Recall & F1 (M=5) &  Precision & Recall & F1 (M=5) \\
       \toprule
       \ourCPD &0.48 (0.08)  &1.00 (0.00) & 0.64 (0.06) &1.00 (0.00)  &1.00 (0.00) & \textbf{1.00 (0.00)} \\
       \ourCPD-CovScore &0.48 (0.07)  &1.00 (0.00) &0.64 (0.06)  &0.39 (0.14)&1.00 (0.00) &0.54 (0.16)\\
       \ourCPD-DistScore &0.00 (0.00)  &0.00 (0.00) &0.00 (0.00)  &1.00 (0.00)&0.96 (0.09) &0.97 (0.06)\\
       \midrule
       KL-CPD & 0.10 (0.10) &0.13 (0.13) &0.11 (0.11)  &0.72 (0.17)&0.56 (0.09) &0.63 (0.12)\\
       Roerich &1.00 (0.00)  &0.98 (0.06) & \textbf{0.99 (0.03)}  &1.00 (0.00)&0.43 (0.12) &0.58 (0.12)\\
      \midrule
    GraphTime  & 0.08 (0.03) &0.95 (0.08) &0.15 (0.06)  &0.62 (0.17) &1.00 (0.00) &0.75 (0.12)\\ 
    TIRE &0.14 (0.13)  &0.10 (0.09) &0.12 (0.11)  &1.00 (0.00) &0.73 (0.21) &0.80 (0.17)\\ 
    \end{tabularx}
    \caption{Performance of CPD methods on the Changing Correlations and Arbitrary CPs with $M=5$.}
    \label{table:ccac5}
\end{table*}

\begin{table*}[ht!]
\small
    \centering
    \begin{tabularx}{\textwidth}{lcccccc} &
    \multicolumn{3}{c}{Bee Dance} &
    \multicolumn{3}{c}{HAR} \\
    \toprule
       Method  &  Precision & Recall & F1(M=5) & Precision & Recall & F1(M=5)\\
       \toprule
       \ourCPD &0.36 (0.18) &0.59 (0.22) & \textbf{0.45 (0.15)} &0.72 (0.06) &0.48 (0.06) & \textbf{0.58 (0.06)} \\
       \ourCPD-CovScore  &0.34 (0.26) &0.36 (0.19) &0.34 (0.21) &0.62 (0.06) &0.56 (0.05) &0.57 (0.04)\\
       \ourCPD-DistScore  &0.48 (0.11) &0.25 (0.13) &0.32 (0.11) &0.50 (0.06) &0.35 (0.03) &0.40 (0.03)\\
       \midrule
       KL-CPD  &0.24 (0.26) &0.10 (0.07) &0.13 (0.11) &0.66 (0.11) &0.20 (0.03) &0.30 (0.04)\\
       Roerich  &0.50 (0.34) &0.32 (0.26) &0.40 (0.30) &0.69 (0.15) &0.11 (0.03) &0.18 (0.05) \\
      \midrule
    GraphTime   &0.13 (0.04) &0.77 (0.13) &0.22 (0.07) &0.04 (0.002) &0.96 (0.02) &0.08 (0.01)\\ 
    TIRE &0.34 (0.44) &0.14 (0.19) &0.20 (0.26) &0.52 (0.19) &0.14 (0.05) &0.22 (0.08)\\
    \end{tabularx}
    \caption{Performance of multiple CPD methods on the Bee Dance dataset.}
    \label{tab:beehac}
\end{table*}

Tables \ref{table:jmcv5}-\ref{tab:beehac} show the performance of TiVaCPD and all baselines on 4 simulated and 2 real-world data sets.
Results are reported as $F_1$ scores, as well as precision and recall, measuring how well it detects CP locations in time. To estimate performance metrics, we define a margin of error $M$ for the exact location of CP, which is common practice in the CPD literature \citep{burg2020, tscpc2}. Given a user-defined margin of error, $M>0$, an estimated CP is a True Positive (TP) if the distance between the ground truth ($t^*$) and the estimated CP ($\widetilde{t}$) is smaller than the margin, i.e. $|t^* - \widetilde{t}|\leq M$. As explained in Figure \ref{fig:evaluations}, if an estimated CP falls outside the margin, then it is considered False Positive (FP), i.e. $\widetilde{t} \notin [t^* - M/2, t^* + M/2]$. We show the impact of margin values $M=\{5,10\}$ on the performance of all baselines in the supplementary material. 


Our method, \ourCPD, effectively detects various types of CPs in simulated datasets by using different components of the score to capture specific CP's types. The CovScore component excels at detecting CPs caused by changes in correlation, the DistScore component detects CPs caused by changes in distribution, and the ensemble score combines the strengths of both components for improved overall performance. Additionally, each component of the score individually outperforms baseline methods, indicating that the performance boost is not just the result of ensembling the scores, but each component of the score itself does a better job at detecting the relevant CPs compared to baseline methods. Although Roerich performs well in detecting simple CPs; however, it fails to address more complex CPs in both synthetic and real-world data. Our method is the only one that performs well in detecting complex CPs. 
The performance results on the real-life datasets are reported in Table \ref{tab:beehac}, for all baselines with a margin value of 5. We show that our method outperforms all baselines in detecting the exact time of CP in real-world and simulation datasets. 
Figure \ref{fig:interactions_real_data} shows a graphical representation and comparison of the different CPD methods for a time-series sample (row 1), and demonstrates how different methods generate scores for CPs. The different components of \ourCPD\ are shown in rows 2 and 3, and show what category of CPs they identify. The heatmaps of CovScore can be used to identify in which pair of features the change in correlation has occurred, and also in which direction the change was.

\section{Discussion}
 In this paper, we introduce \ourCPD, a novel CPD method  for detecting and characterizing various types of CPs in time-series data. By capturing changes in feature distribution, dynamics, and correlation networks, \ourCPD \space provides valuable insights into the underlying causes of CPs, enhancing interpretability for end users. This is particularly crucial in domains like healthcare, where the type of CP significantly influences downstream decision making.
The method is currently designed for offline settings to retrospectively detect changes. For future work, we intend to extend \ourCPD\ to the online setting, where real-time measurements are acquired.
Moreover, we plan to incorporate techniques for imputing missing data by leveraging correlated features and temporal dynamics.
 


\bibliography{references}
\bibliographystyle{bibstyle}


\end{document}